\documentclass[a4paper,fleqn]{cas-dc}

\usepackage[numbers]{natbib}
\usepackage{tabularx}

\usepackage{multirow}
\usepackage{makecell}
\usepackage{booktabs}
\usepackage{tabularx}

\def\tsc#1{\csdef{#1}{\textsc{\lowercase{#1}}\xspace}}
\tsc{WGM}
\tsc{QE}
\tsc{EP}
\tsc{PMS}
\tsc{BEC}
\tsc{DE}

\begin{document}
\let\WriteBookmarks\relax
\def\floatpagepagefraction{1}
\def\textpagefraction{.001}

\shorttitle{Bengali MCQ Generation and Answer Prediction}
\shortauthors{Surzo et~al.}

\title[mode = title]{BengaliMCQ: Automatic Generation and Answer Prediction of Academic Multiple-Choice Questions in a Low-Resource Language}

\author[a]{Abu Tarabin Surzo}[orcid=https://orcid.org/0009-0002-1099-7337]
\ead{abu.tarabin.surzo@g.bracu.ac.bd}
\cormark[1]

\credit{Conceptualization, Methodology, Writing – original draft}

\author[a]{A.K.M. Nihalul Kabir}[orcid=https://orcid.org/0009-0002-9939-6553]
\ead{akm.nihalul.kabir@g.bracu.ac.bd}

\credit{Methodology,  Writing – original draft, Writing – review and editing}

\author[a]{Sm Azmain Faysal}[orcid=https://orcid.org/0009-0003-1966-7729]
\ead{sm.azmain.faysal@g.bracu.ac.bd}

\credit{Methodology, Formal analysis, Visualization}

\author[a]{Ariana Haque Ami}
\ead{ariana.haque.ami@g.bracu.ac.bd}

\credit{Data curation, Formal analysis}

\author[a]{Lawrence Amlan Gomes}
\ead{lawrence.amlan.gomes@g.bracu.ac.bd}
\credit{Data curation, Investigation}

\author[a]{Farig Sadeque}[orcid=https://orcid.org/0000-0001-6797-7826]
\ead{farig.sadeque@bracu.ac.bd}
\credit{Conceptualization, Supervision}

\affiliation[a]{%
    organization={Department of Computer Science and Engineering, BRAC University},
    addressline={Kha 224 Pragati Sarani, Merul Badda},
    city={Dhaka},
    postcode={1212},
    country={Bangladesh}
}

\cortext[cor1]{Corresponding author}

\begin{abstract}
Traditional retrieval-augmented generation (RAG) frameworks process documents without attending to their hierarchical structure, leading to poor performance, especially in low-resource languages such as Bengali. To address this, we propose a structure-aware RAG framework that models Bengali textbooks as hierarchical graphs and uses a contrastively trained graph neural network to retrieve a small set of relevant passages. These passages provide focused context for a large language model, enabling topic-specific multiple-choice question (MCQ) generation and in-domain answer prediction. Experimental results demonstrate that our framework outperforms strong dense retrieval baselines across retrieval metrics, produces more relevant MCQs, and achieves superior answer prediction accuracy.
\end{abstract}




\maketitle

\section{Introduction}\label{sec:intro}

Automatic MCQ generation has become a popular downstream application of large language models (LLMs), especially with recent developments in their capabilities. However, their effectiveness depends heavily on the relevance of the input context provided to them. For topic-specific question generation, it is impractical to feed entire textbook chapters as input to LLMs due to their context-window limits. Even when an entire chapter is provided, the language models may not consistently attend to the most relevant passages. Retrieval‑augmented generation (RAG) addresses this by providing the LLMs with a comparatively smaller, targeted set of context for generation. Standard dense retrieval often ignores the hierarchical structure of documents, resulting in a negative impact on performance, especially if the text is in a low-resource language like Bengali. 

Graph representations can address this limitation by capturing various relationships within documents. Most existing work, however, focuses on high-resource languages. In this paper, we introduce BengaliMCQ, a document-graph-guided automated system that generates highly topic-relevant academic MCQs in Bengali and predicts answers to in-domain MCQs with high accuracy. To summarize, our contributions are:

\begin{itemize} 

\item We implement a GNN-based retrieval framework that learns to rank query-relevant passages by modeling a textbook’s hierarchical structure, allowing us to reduce context length without sacrificing coverage.

\item We evaluate our approach through automatic metrics as well as expert validation, showing that our framework outperforms baselines in retrieval quality, question quality, and answer accuracy. 

\item We provide an ablation study for analyzing the contribution of our design choices and demonstrating their value.

\end{itemize} 

\section{Related work}
Though early work on automatic MCQ generation depended on probabilistic and rule-based approaches like TF-IDF, n-grams, and part-of-speech tagging \citep{A_Nwafor_2021}, this field soon progressed to neural-network and transformer-based frameworks. For instance, fine-tuning T5- and BERT-based models \citep{Roy, Patil2022} has become the de facto standard for downstream NLP tasks like question generation, named-entity recognition, and question answering. A notable hybrid method incorporating transformers with a rule-based approach was proposed by \citet{Mehta2021AutomatedMG}, which utilized the BERTSUM model to summarize text and the Rapid Automatic Keyword Extraction algorithm to extract keywords for creating fill-in-the-blank MCQs. Distractors (incorrect options) were generated using WordNet based on hypernym and hyponym relationships. However, distractors of higher quality can be generated by LLMs through structured multi-stage prompting, as shown by \citet{maity}. Currently, this field is being dominated by LLMs. Recent works have tried to leverage LLMs for automatic MCQ generation and evaluation by utilizing different prompt engineering techniques combined with human-guided reviews \citep{mucciaccia-etal-2025-automatic}. However, these systems typically suffer from performance degradation on long-context documents \citep{hsieh2024ruler}. To the best of our knowledge, no prior work exploits the document structure of textbooks for generating topic-relevant MCQs. 


Recent research has begun to fuse graphs with RAG for mapping and modeling these structural relationships. \citet{iyer-etal-2023-question} trained a GNN on a temporary graph (during offline learning) created by modeling relationships between question–answer sentence pairs for answer sentence selection (AS2) tasks. Another recent GNN-based framework closely related to ours is AutoRev \citep{chitale2026graphguidedpassageretrievalauthorcentric}, which models the structure of research papers as hierarchical documents. The authors used a GNN to retrieve the most salient passages to automatically generate academic peer reviews. Inspired by these approaches, our work employs GNN-based contextual learning to retrieve the most topic-relevant passages for MCQ generation and answer prediction.

\section{Methodology}\label{sec:methodology}

\subsection{Document processing} \label{document_processing}
As target books, we chose four secondary-school-level Bengali textbooks: "Bangla Shahitto" and "Shohopath" from the literature domain, "Biology" from the STEM domain, and "Bangladesh and Global Studies" (BGS) from the social science domain. Using \textit{Tesseract-4} optical character recognition \citep{Smith2007AnOO} and Google Lens, we extracted and parsed the text to detect chapters, headings, subheadings, passages, and sentences for the node creation in Section \ref{subsec:graph-construction}. The original passage structure of the books was preserved by splitting on newline characters to maintain the inherent cohesion of the academic content. 

\subsection{Query-Passage dataset creation}\label{subsec:query-passage}
As our goal was to train a GNN for ranking relevant passages, we required a labeled dataset where queries (comprising topic sentences and short questions) would be mapped to their relevant passages. However, due to the lack of such a dataset tailored to our books, we created a synthetic one utilizing state-of-the-art LLMs such as \textit{Gemini 3.1 Pro} and \textit{GPT-4}. We prompted one model to generate and another to curate diverse queries from individual passages or small groups of passages. Our final dataset contains around 22,000 query-passage pairs with roughly 5,000-6,000 from each book. For the training-validation split, we use an 80:20 ratio.



\begin{figure}
	\centering
	\includegraphics[width=\columnwidth]{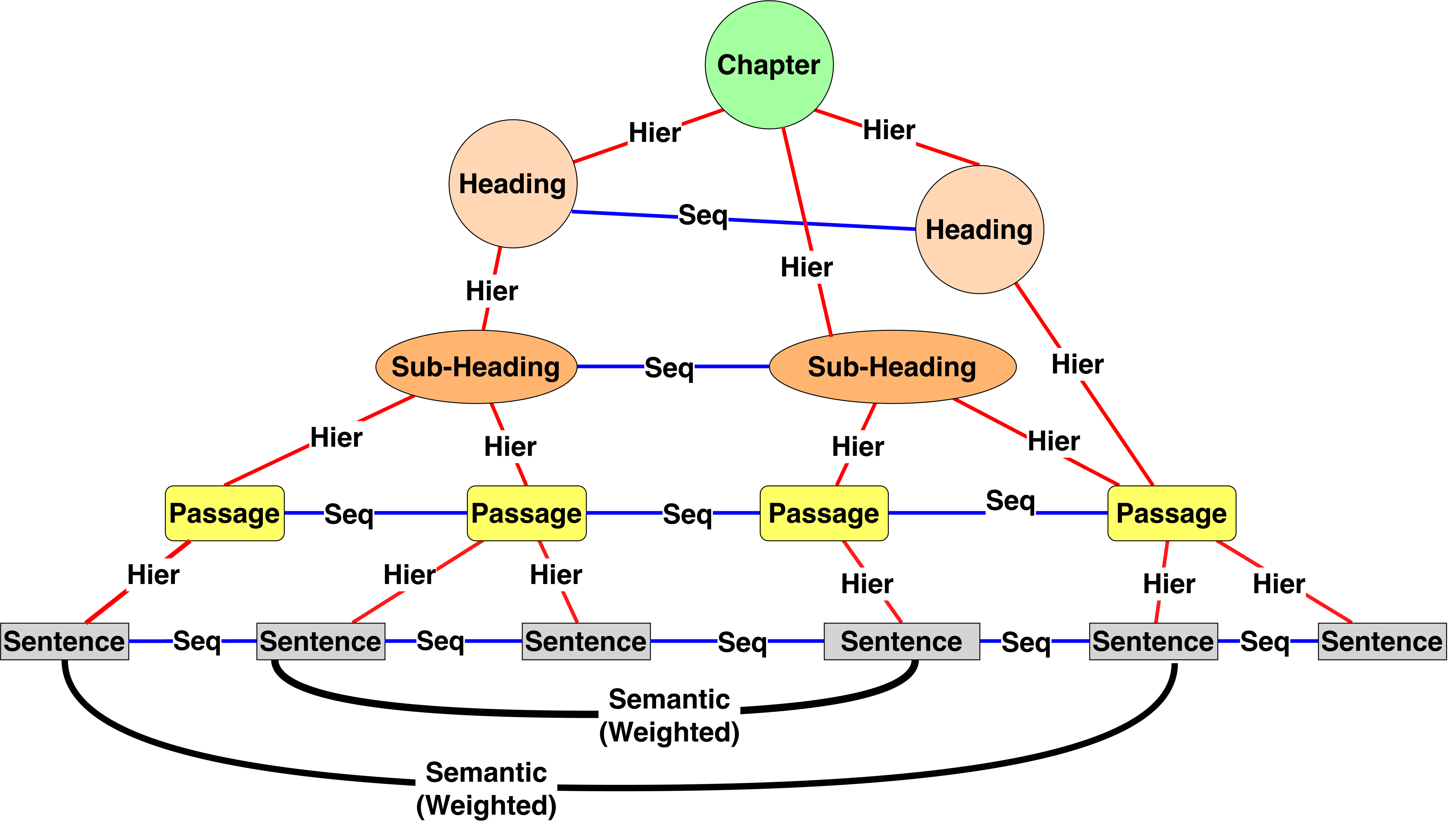}
	\caption{A simple schematic illustration of our textbook's hierarchical graph. Seq = Sequential edge, Hier = Hierarchical edge}
	\label{FIG:1}
\end{figure}

\subsection{Graph construction}\label{subsec:graph-construction}
We borrow and enrich the graph representation suggested by \citet{chitale2026graphguidedpassageretrievalauthorcentric} by adding weighted semantic edges for better contextualization. Let each book be represented as a hierarchical directed multigraph ($\mathcal{G} = (\mathcal{V}, \mathcal{E})$), where nodes $\mathcal{V}$ correspond to the parsed chapters, headings, subheadings, passages, and sentences. Subsequently, we add bidirectional hierarchical ($\mathcal{E_{\text{hier}}}$) and sequential edges ($\mathcal{E_{\text{seq}}}$) between the nodes to allow message-passing both throughout the document hierarchy and along the sequential flow of the book. 

The representation is further enhanced by adding weighted semantic edges between sentence nodes to capture long-range relationships between related content. For each sentence $s$, we add bidirectional semantic edges ($\mathcal{E_{\text{sem}}}$) to its top-{$K_{s}$} ($K_s \in \{3, 5\}$) most similar neighbors whose cosine similarity exceeds a threshold {$T$} ($T \in \{0.5, 0.6, 0.7\}$). The similarity scores are set as the edge weights. For obtaining the embeddings, we use a multilingual bi-encoder, namely \textit{BGE-M3} \citep{chen2025m3embeddingmultilingualitymultifunctionalitymultigranularity}, fine-tuned on our query-passage dataset. A simple illustration of our document graph for a book is shown in Figure \ref{FIG:1}.

\subsection{GNN architecture and training}
For cohesively attending to several types of edges in our graphs, we perform supervised training of a graph attention network (GAT) using the dataset created in Section~\ref{subsec:query-passage}. Node features are initialized with embeddings ($dim=1024$) generated by our fine-tuned bi-encoder. For the passage nodes, we
also evaluated initialization with zero vectors and with the mean-pooled embedding of child nodes. 


\subsubsection{Weighted graph attention layer}
Our graphs are heterogeneous, with a combination of weighted and unweighted edges; we therefore design a custom multi-head attention layer. We integrate the semantic edge-weight \(w_{ij}\) with the attention score through a learnable scalar \(\lambda \) and add an edge-type-specific learnable bias parameter \(b^{type}\). Thus, for an edge from node \(i\) to node \(j\) the base unnormalized attention score becomes:

\vspace{-10pt}
\begin{equation}
\label{eqn1}
e_{ij} =
\bigl[x'_i \parallel x'_j\bigr]\cdot a
+\lambda\cdot w_{ij}
+b^{type},
\end{equation}
\vspace{-10pt}

where \(a \in \mathbb{R}^{\mathcal{{H \times D}}}\) ($\mathcal{H} = $ Attention heads, $D =$ Output dimension per head) is a learnable attention vector; ${{x}_i'}$ and ${{x}_j'}$ (\({{x}_i'}, {{x}_j'} \in \mathbb{R}^{\mathcal{{H \times D}}}\)) denote the linear projections of the initial node embeddings ${{x}_i}$ and ${{x}_j}$ (\({{x}_i}, {{x}_j} \in \mathbb{R}^{dim}\)). The score in Eqn. \ref{eqn1} is then fed through a standard LeakyReLU and softmax normalization for computing the final attention coefficient $\alpha_{ij}$. 


Aggregation is performed by summation, weighted by the attention coefficients. Our pipeline contains three such attention layers stacked sequentially. Their outputs are concatenated with the original node embedding and then projected back down to the initial 1024 dimensions via a lightweight fusion multi-layer perceptron (MLP) to obtain the learned node embedding \(\mathit{h}_i\). This residual design mitigates the typical oversmoothing problem of GNNs \citep{rusch2023surveyoversmoothinggraphneural} and preserves the initial signal.


\subsubsection{Training objective} 
We use a contrastive learning objective, specifically the multi-positive InfoNCE loss to train the GAT as a passage ranking model. The loss function we use is:



\vspace{-5pt}
\begin{equation}
\label{infonce}
\mathcal{L}
=
-\log\left(
\frac{
\displaystyle\sum_{i\in P^{+}}
\exp\left(\tau\cdot\operatorname{cossim}
(q,h^{passage}_i)\right)
}{
\displaystyle\sum_{x=1}^{N_b}
\exp\left(\tau\cdot\operatorname{cossim}
(q,h^{passage}_x)\right)
}
\right)
\end{equation}
\vspace{-5pt}

where $q$ denotes the query embedding, $P^{+}$ denotes its ground-truth positive passages, \(\tau>0\) is a learnable temperature parameter, ${h}^{passage}_i$ is the GAT-learned embedding of passage node $i$, and $N_{{b}}$ is the total number of passages in the book. The loss in Eqn. \ref{infonce} trains the model to assign high scores to all positive (relevant) passages while suppressing the scores of all other passages of that book. The GAT is trained using various configurations based on the values of $K_{s}$ and $T$, as described in the semantic edge construction process in Section \ref{subsec:graph-construction}. These trained GATs are denoted with \({GAT} (K_s, T)\).

\begin{figure}
	\centering
	\includegraphics[width=\columnwidth]{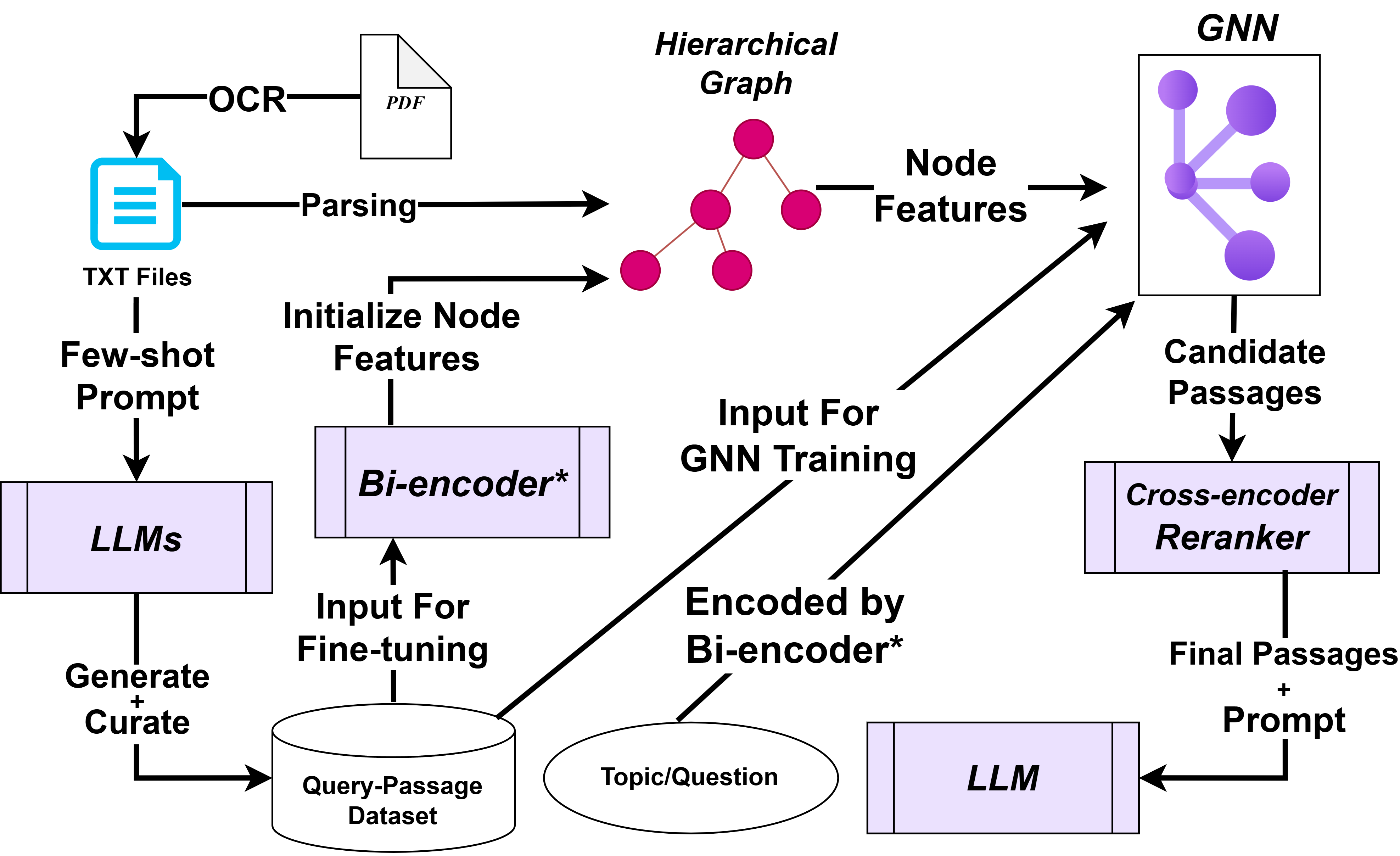}
	\caption{End-to-end BengaliMCQ framework}
	\label{FIG:2}
\vspace{-8pt}
\end{figure}

\begin{figure}
	\centering
	\includegraphics[width=0.7\columnwidth, height=7cm, keepaspectratio=false]{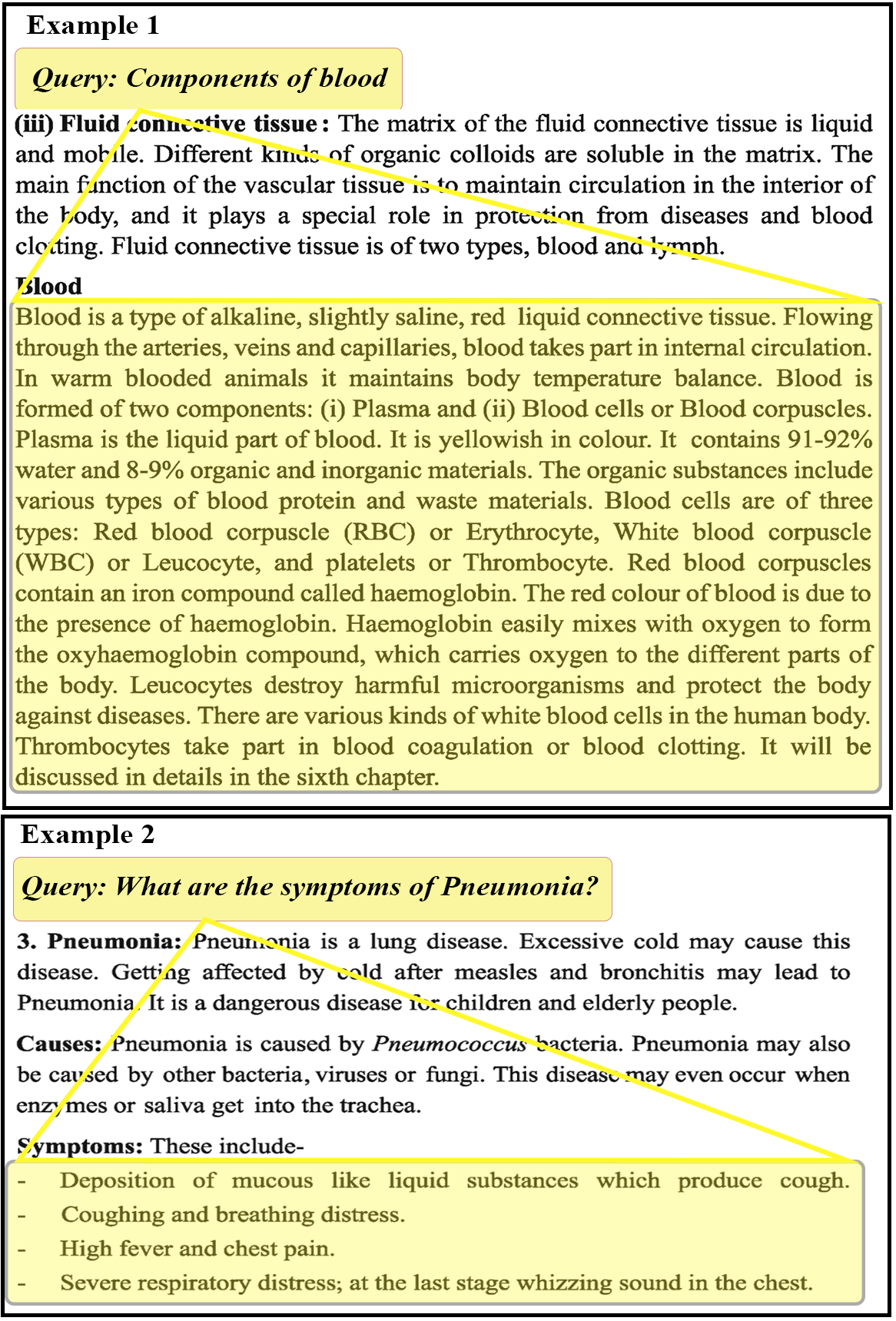}
	\caption{ Excerpt of two test samples where passages highlighted in yellow color refer to query-relevant passages identified GNN but missed by DPR}
	\label{FIG:3}
\vspace{-7pt}
\end{figure}
\subsection{Passage retrieval and grounding}
During inference, a user inputs a topic for MCQ generation or a question for answer prediction. The system then uses the trained GAT's ranking to retrieve the top-$K_c$ ($K_c=10$) candidate passages that are relevant to the input topic or question. We perform a further re-ranking on these with a cross-encoder, namely \textit{BGE-Reranker-V2-M3} \citep{chen2025m3embeddingmultilingualitymultifunctionalitymultigranularity}, for improved performance and smaller context. Finally, the top-$K_f$ (with $K_f = 5$) most relevant passages after the two-stage ranking are taken as the grounded context. Augmented with a structured prompt, these passages are sent to the LLM as input for topic-focused question generation or answer prediction of MCQs. Selecting the small set of pertinent passages for a given topic or MCQ helps reduce the context window and computational cost. An illustration of each step of the entire framework is shown in Figure \ref{FIG:2}.

\section{Results and analysis}\label{sec:results_analysis}

This section presents a comprehensive evaluation of our framework along three aspects: retrieval performance, downstream task performance, and ablation studies of our design choices. 

\begin{table*}[t]
\centering
\caption{Retrieval performance comparison across different methods. Bold highlights the best scores, underlined values indicate second-best. }
\label{tab:retrieval_performance}
\small
\setlength{\tabcolsep}{4pt}
\begin{tabularx}{\textwidth}{
    >{\raggedright\arraybackslash}X
    *{8}{>{\centering\arraybackslash}p{0.075\textwidth}}
}
\toprule
\textbf{Baseline} &
\textbf{HR@5} &
\textbf{Recall@5} &
\textbf{nDCG@5} &
\textbf{MRR@5} &
\textbf{HR@10} &
\textbf{Recall@10} &
\textbf{nDCG@10} &
\textbf{MRR@10} \\
\midrule
BM25 & 0.539 & 0.506 & 0.506 & 0.506 & 0.600 & 0.600 & 0.453 & 0.434 \\
Translation Baseline (DPR) & 0.753 & 0.710 & 0.618 & 0.608 & 0.821 & 0.782 & 0.642 & 0.617 \\
Sparse Retrieval (\textit{BGE-M3}) & 0.795 & 0.760 & 0.676 & 0.666 & 0.852 & 0.820 & 0.696 & 0.674 \\
DPR (\textit{Harrier-OSS-v1}) & 0.519 & 0.487 & 0.418 & 0.411 & 0.575 & 0.545 & 0.438 & 0.419 \\
DPR (\textit{Harrier-OSS} Fine-tuned) & 0.836 & 0.802 & 0.703 & 0.689 & \underline{0.901} & \underline{0.873} & 0.728 & 0.698 \\
DPR (\textit{BGE-M3} Fine-tuned) & \underline{0.842} & \underline{0.803} & \underline{0.707} & \underline{0.696} & 0.899 & 0.866 & \underline{0.729} & \underline{0.703} \\
GAT($K_s = 5$, $T = 0.5$) & \textbf{0.882} & \textbf{0.851} & \textbf{0.740} & \textbf{0.720} & \textbf{0.934} & \textbf{0.911} & \textbf{0.761} & \textbf{0.726} \\
\midrule
BGE-M3+RERANKER & \underline{0.903} & \underline{0.871} & \underline{0.815} & \underline{0.814} & \underline{0.929} & \underline{0.901} & \underline{0.826} & \underline{0.818} \\
GAT+RERANKER & \textbf{0.922} & \textbf{0.886} & \textbf{0.828} & \textbf{0.817} &
\textbf{0.954} & \textbf{0.922} & \textbf{0.842} & \textbf{0.831} \\
\bottomrule
\end{tabularx}
\vspace{-5pt}
\end{table*}

\subsection{Retrieval performance}

We evaluate the retrieval quality of the proposed framework by comparing against relevant baselines on the validation split (approximately 4400 samples) of the query-passage dataset. The baselines include: BM25, lexical-weight-based sparse retrieval, several dense passage retrieval (DPR) pipelines, DPR + cross-encoder re-ranking hybrids, and a translation-based DPR pipeline. Following prior retrieval work, we report normalized discounted cumulative gain (nDCG@k), mean reciprocal rank (MRR), recall@k and hit rate (HR@k) per method. Higher values indicate better retrieval performance.

 As shown in Table \ref{tab:retrieval_performance}, our graph attention network consistently outperforms all baselines that do not use a re-ranker across the retrieval metrics. Notably, when comparing re-ranker integrated baselines, the nDCG@5 obtained with only 5 passages by our "\textit{GAT + BGE-Reranker}" pipeline surpasses the nDCG@10 obtained with 10 passages by a fine-tuned "\textit{BGE-M3 + BGE-Reranker}" baseline, demonstrating that our framework achieves better performance with less context. Figure \ref{FIG:3} shows empirical examples where GAT was able to retrieve passages that other baselines missed.

\subsection{End-to-end evaluation}

\begin{table*}[t]
\centering
\caption{Comparison of MCQ generation and answer prediction performance across different baselines. All metrics are reported as averages across all books for each baseline.}
\label{tab:mcq_generation_evaluation}
\small
\setlength{\tabcolsep}{3pt}
\begin{tabularx}{\textwidth}{
    >{\raggedright\arraybackslash}X                      
    *{6}{>{\centering\arraybackslash}p{0.07\textwidth}}  
    >{\centering\arraybackslash}p{0.14\textwidth}        
    >{\centering\arraybackslash}p{0.14\textwidth}        
}
\toprule
\textbf{Baseline} &
\textbf{IWF} &
\textbf{QF} &
\textbf{TR} &
\textbf{DS} &
\textbf{PPL} &
\textbf{Div} &
\textbf{\makecell{Answ. \\ Accuracy (\%)}} &
\textbf{\makecell{Aver. Context \\ Length (Tokens)}} \\
\midrule

Zero-shot & 0.39 & \underline{14.53} & 0.46 & 0.71 & \underline{7.60} & 0.44 & 57.21 & 43,591 \\
Few-shot & 0.31 & 14.40 & 0.46 & \underline{0.77} & 7.83 & 0.45 & 68.35 & 43,591 \\
CoT & \textbf{0.22} & 14.50 & 0.48 & \textbf{0.78} & 7.98 & \textbf{0.58} & 71.87 & 43,591 \\
DPR + Reranker & 0.27 & 13.96 & \underline{0.59} & 0.75 & 7.88 & \underline{0.52} & \underline{82.80} & \underline{2,825} \\
GAT + Reranker & \underline{0.24} & \textbf{14.81} & \textbf{0.67} & \textbf{0.78} & \textbf{7.57} & \underline{0.52} & \textbf{91.41} & \textbf{1,651} \\

\bottomrule
\end{tabularx}
\end{table*}

To assess downstream performance, we have evaluated our framework against long-context zero-shot, few-shot, and chain-of-thought (CoT) prompting baselines, and the hybrid retrieval strategy of \textit{DPR + Re-ranker}. For each method, the same LLM (\textit{Gemini 3.1}) is used for question generation and answer prediction.

\subsubsection{MCQ generation evaluation protocol} 
For MCQ evaluation, 10 topics from each book are selected and curated by our domain experts to ensure that our books contain sufficient material on these topics. For every topic, each baseline generates 20 multiple-choice questions, resulting in a total of 800 MCQs per method. The generated MCQs are then evaluated using two complementary approaches. First, we have used the "LLM as a judge" technique by prompting GPT-5.2 with a structured Item-Writing Flaws (IWF) rubric \citep{10.1007/978-3-031-42682-7_16} and the QUEST framework (QF) rubric \citep{10.1007/978-3-031-95627-0_20}. The IWF rubric considers the pedagogical value of a question and its options through various criteria. The QF rubric, on the other hand, evaluates MCQs across five dimensions: Quality, Uniqueness, Effort, Structure, and Transparency. Second, we report automatic metrics that quantify different aspects of question quality: 
\paragraph{Distractor similarity (DS)} 
To evaluate the plausibility effect of the distractors, we have used the mean-pooled embeddings of a fine-tuned BanglaBERT \citep{DBLP:journals/corr/abs-2101-00204} model. For each MCQ, the cosine similarities between the embedding of the correct answer and that of the distractors are calculated. The model was fine-tuned on our books using masked language modeling (MLM). 

\paragraph{Topic relevance (TR)} 
To measure how relevant our LLM-generated MCQs are to the topic sentences, we also compute the cosine similarity between the sentence embedding of each question and its corresponding topic sentence using a second multilingual sentence encoder, \textit{Harrier-OSS-v1-0.6b}. 

\paragraph{Perplexity of questions (PPL)}
To quantify the sentence formation quality, we report the mean perplexity of the question tokens under a lightweight fine-tuned generative model for Bengali, shahidul034/BanglaGPT. Similar to BanglaBERT, this was also fine-tuned on our textbooks but using causal language modeling (CLM).  

\paragraph{Question diversity (Div)}
Question type diversity was evaluated using Shannon entropy, following \citet{Raina}. For finding the question categories, a generative LLM was few-shot prompted to classify each MCQ into one of the 10 question types: What, Who, When, Where, Why, How, Which, Yes/No, Whose, How much/many.

Higher values of QF, DS, TR, and Div indicate better performance, whereas lower values of IWF and PPL are preferred. Table \ref{tab:mcq_generation_evaluation} summarizes the quantitative results.

\subsubsection{Answer prediction evaluation}
Unlike MCQ generation, answer prediction performance can be measured with a simple metric like accuracy. The assessment was done on a held-out dataset of 1,000 gold-standard MCQs collected from past secondary school public exam archives. We manually curated this dataset to ensure that the questions correspond to our target textbooks. The results in Table \ref{tab:mcq_generation_evaluation} indicate that our system achieves an average accuracy of 91.41\%, which is the highest out of all baselines. These findings indicate that
our model generalizes across the academic text types in our corpus.

\subsection{Ablation study}
We conduct ablation experiments to investigate the retrieval impact of key design choices in the proposed framework. All experiments are performed on the same validation split, with \(K_s\) = 5 and \(T\) = 0.7 unless otherwise stated. 

\begin{table*}[t]
\centering
\caption{Ablation study of the proposed GAT-based retrieval framework. The “All Edges”, “Zero” and “Using Temp” rows report the same reference configuration and are therefore
identical.}
\label{tab:ablation}
\small
\setlength{\tabcolsep}{4pt}
\begin{tabularx}{\textwidth}{
    >{\centering\arraybackslash}p{0.14\textwidth}
    >{\raggedright\arraybackslash}X
    *{8}{>{\centering\arraybackslash}p{0.075\textwidth}}
}
\toprule
\textbf{Dimension} &
\textbf{Baseline} &
\textbf{HR@5} &
\textbf{Recall@5} &
\textbf{nDCG@5} &
\textbf{MRR@5} &
\textbf{HR@10} &
\textbf{Recall@10} &
\textbf{nDCG@10} &
\textbf{MRR@10} \\
\midrule

\multirow{4}{*}{\makecell{\textbf{Edge}\\\textbf{Type}}}
& No Hier 
& 0.013 & 0.012 & 0.007 & 0.006 & 0.016 & 0.014 & 0.008 & 0.007 \\

& No Seq 
& \underline{0.865} & 0.824 & 0.724 & 0.712 
& \textbf{0.920} & \underline{0.887} & 0.745 & 0.719 \\

& No Sem 
& \underline{0.865} & \underline{0.826} & \underline{0.727} & \textbf{0.716}
& \textbf{0.920} & 0.886 & \underline{0.748} & \textbf{0.723} \\

& All Edges 
& \textbf{0.868} & \textbf{0.829} & \textbf{0.728} & \textbf{0.716}
& \textbf{0.920} & \textbf{0.889} & \textbf{0.749} & \textbf{0.723} \\

\midrule

\multirow{3}{*}{\makecell{\textbf{Passage}\\\textbf{Initialization}}}
& MoS 
& 0.857 & 0.818 & 0.711 & 0.697 & 0.915 & 0.882 & 0.734 & 0.704 \\

& Zero 
& \underline{0.868} & \underline{0.829} & \underline{0.728} & \underline{0.716}
& \underline{0.920} & \underline{0.889} & \underline{0.749} & \underline{0.723} \\

& Bi-encoder 
& \textbf{0.879} & \textbf{0.839} & \textbf{0.733} & \textbf{0.719}
& \textbf{0.929} & \textbf{0.897} & \textbf{0.753} & \textbf{0.725} \\

\midrule

\multirow{2}{*}{\makecell{\textbf{Temperature}\\\textbf{Parameter}}}
& No Temp 
& 0.819 & 0.778 & 0.719 & 0.696 & 0.846 & 0.806 & 0.729 & 0.703 \\

& Using Temp 
& \textbf{0.868} & \textbf{0.829} & \textbf{0.728} & \textbf{0.716}
& \textbf{0.920} & \textbf{0.889} & \textbf{0.749} & \textbf{0.723} \\

\midrule

\multirow{6}{*}{\makecell{\textbf{GAT($K_s$, $T$)}\\\textbf{Sensitivity}}}
& GAT(3, 0.5) 
& 0.878 & 0.839 & 0.735 & \underline{0.721}
& \underline{0.932} & 0.901 & 0.756 & \underline{0.728} \\

& GAT(3, 0.6) 
& 0.878 & 0.839 & \underline{0.737} & \textbf{0.724}
& 0.930 & 0.899 & \underline{0.758} & \textbf{0.731} \\

& GAT(3, 0.7) 
& 0.876 & 0.837 & 0.733 & 0.719
& 0.930 & 0.898 & 0.754 & 0.727 \\

& GAT(5, 0.5) 
& \textbf{0.882} & \textbf{0.851} & \textbf{0.740} & 0.720
& \textbf{0.934} & \textbf{0.911} & \textbf{0.761} & 0.726 \\

& GAT(5, 0.6) 
& \underline{0.880} & \underline{0.842} & \textbf{0.739} & \textbf{0.724}
& 0.929 & \underline{0.897} & \underline{0.758} & \textbf{0.731} \\

& GAT(5, 0.7) 
& 0.875 & 0.836 & 0.732 & 0.718
& \underline{0.932} & 0.900 & 0.754 & 0.725 \\

\bottomrule
\end{tabularx}
\vspace{-5pt}
\end{table*}

\paragraph{Effect of edge types}
We discard each type of edge while keeping the rest of the architecture fixed. In this case, all passage nodes are initialized with zero vectors (\(dim = 1024\)) to make sure that the ranking signal is only provided by the message-passing on the graph. As seen in Table \ref{tab:ablation}, removing hierarchical edges causes performance to collapse, indicating that hierarchical structure is the most important signal to be learned in the textbook graph.  By contrast, the drop from discarding sequential or semantic edges is measurable but modest relative to the full model. 

\paragraph{Effect of passage-node initialization}
We compare the following three initialization methods for passage nodes: (i) mean-pooling of child sentence embeddings, (ii) directly encoding with the fine-tuned \textit{BGE-M3} encoder, and (iii) initialization with zero vectors. Interestingly, zero initialization outperforms mean-pooling, although direct encoding with \textbf{BGE-M3} achieves the best overall performance.

\paragraph{Effect of temperature parameter  \(\tau>0\) }
From our experimental observations, training with a learnable temperature produces better ranking metrics than a fixed temperature of 1.0 (equivalent to not using temperature). The learned value typically settles between 15 and 20.

\paragraph{Effect of hard negative mining}
We tested offline hard negative mining (TopK-MarginPos) \citep{moreira2025nvretrieverimprovingtextembedding} with the same fine-tuned BGE-M3 bi-encoder. Our experiments show that hard
negative mining substantially degraded validation performance (Recall@5 below 50\%) and prevented our model from learning effectively. 


\paragraph{Impact of varying $K_s$ and $T$ values}
We test varying top-$K_s$ and threshold $T$ values for semantic edge creation and report the impact on retrieval performance. From Table \ref{tab:ablation}, we can see that although the performance impact varies depending on the metric, ($K_s = 5, \ T=0.5$) yields the best retrieval performance. Therefore, we adopted these values for the full re-ranker-integrated pipeline.




\subsection{Expert validation}

\begin{table}[htbp]
\centering
\footnotesize 
\setlength{\tabcolsep}{4pt} 
\caption{Human evaluation of topic relevance (1 = not relevant, 2 = somewhat relevant, 3 = relevant, and 4 = highly relevant).}
\label{tab:human_eval}
\begin{tabular}{@{}lccc@{}}
\toprule
\textbf{Domain} & \textbf{$N$} & \textbf{Mean} & \textbf{\% $\ge 3$} \\ \midrule
Literature         & 60           & 3.17          & 76.7               \\
Biology            & 40           & 3.58          & 92.5               \\
Social Science$^*$ & 40           & 3.13          & 80.0               \\ \midrule
\textbf{Overall}   & \textbf{140} & \textbf{3.27} & \textbf{82.1}      \\ \bottomrule
\multicolumn{4}{@{}l}{\scriptsize $^*$Averaged across both experts ($\kappa = 0.55$).}
\end{tabular}
\vspace{-10pt}
\end{table}

We were aware of the limitations of automatic metrics and therefore conducted a small-scale expert validation of our generated MCQs' topic relevance on a 4-point scale. Two domain experts (literature and STEM) evaluated 140 randomly sampled MCQs distributed as follows: 60 MCQs from the literature textbooks ("Shohopath" and "Bangla Shahitto"), 40 from biology, and 40 from the social science (BGS) textbook. The results (Table \ref{tab:human_eval}) demonstrate that 82.1\% of all MCQs are scored as relevant (score $\geq 3$), with biology MCQs performing the best. For estimating inter-annotator agreement, both experts independently rated the 40 Social Science MCQs. Their evaluations yielded moderate agreement (Cohen’s $\kappa = 0.55$), with both annotators independently rating 80.0\% of the cross-domain questions as relevant (score $\geq 3$).



\section{Conclusion}\label{sec:conclusions}
In this work, we develop BengaliMCQ, an automated system for topic-relevant MCQ generation and answer prediction for a low-resource language. By modeling textbooks as hierarchical document graphs and leveraging a GNN to retrieve topic-relevant content, BengaliMCQ reduces LLM input length while maintaining high coverage. Our experiments demonstrate strong empirical performance through automatic metrics and expert validation, reducing the average context length from 43,591 to 1,651 tokens while raising answer prediction accuracy to 91.41\%. The strategies provided in this paper can help close the gap between structure-aware retrieval and automated educational technology. 

A central limitation of our work is that our GNN training is supervised by a synthetic dataset, which may introduce noise due to hallucination \citep{tan-etal-2024-large}. A completely human-curated dataset would likely provide a more reliable and robust training signal. Another limitation lies in our automatic evaluation process, as LLM- and embedding-based scores are only approximations. While useful, our human evaluation was limited to a small stratified subset of topics and items, as well as to only two domain experts. In our future work, we plan to conduct broader, multi-annotator studies with a larger sample of books as well as a human-curated dataset for reliable supervision.



\section*{Acknowledgements}
The authors would like to express their sincere gratitude to Abul Kalam Azad and Azmal Hossain from Ideal School and College, Motijheel, Dhaka, for their valuable time, insights, and contributions during the expert validation of the generated multiple-choice questions (MCQs).

\printcredits

\bibliographystyle{model1-num-names}


\end{document}